\documentclass[]{ceurart}
\usepackage{listings} %

\usepackage{tabularx}
\usepackage{nth}
\usepackage{array}
\usepackage{multirow}
\usepackage{subcaption}
\usepackage{todonotes}
\usepackage{amsfonts}
\usepackage{tcolorbox}
\usepackage{bookmark}
\tcbuselibrary{skins,theorems,breakable}
\usepackage{soul}
\DeclareRobustCommand{\hlgray}[1]{{\sethlcolor{lightgray!50}\hl{#1}}}

\usepackage{arydshln}
\usepackage{adjustbox}
\usepackage{xurl}

\begin{document}

\copyrightyear{2025}
\copyrightclause{Copyright for this paper by its authors.
  Use permitted under Creative Commons License Attribution 4.0
  International (CC BY 4.0).}

\conference{2nd International Workshop on {AI} in Society, Education and Educational Research (AISEER),
  October 25--30, 2025, Bologna, Italy} %

\title{{NLP} Methods May Actually Be Better Than Professors at Estimating Question Difficulty}

\author[1,3]{Leonidas Zotos}[%
  email=l.zotos@rug.nl
]
\cormark[1] %

\author[2]{Ivo Pascal de Jong}[%
  email=ivo.de.jong@rug.nl
]
\cormark[1] %

\author[2]{Matias Valdenegro-Toro}[]

\author[2]{Andreea Ioana Sburlea}[]

\author[1]{Malvina Nissim}[]

\author[3]{Hedderik van Rijn}[]

\address[1]{Center for Language and Cognition, University of Groningen}
\address[2]{Bernoulli Institute, University of Groningen}
\address[3]{Department of Experimental Psychology, University of Groningen}

\cortext[1]{Corresponding authors; These authors contributed equally to this work.}

\begin{abstract}
Estimating the difficulty of exam questions is essential for developing good exams, but professors are not always good at this task. We compare various Large Language Model-based methods with four professors in their ability to estimate what percentage of students will give correct answers on True/False exam questions in the areas of Neural Networks and Machine Learning. Our results show that the professors have limited ability to distinguish between easy and difficult questions and that they are outperformed by directly asking \texttt{Gemini 2.5} to solve this task. Yet, we obtained even better results using uncertainties of the LLMs solving the questions in a supervised learning setting, using only $42$ training samples. We conclude that supervised learning using {LLM} uncertainty can help professors better estimate the difficulty of exam questions, improving the quality of assessment.
\end{abstract}

\begin{keywords}
  item difficulty estimation \sep
  uncertainty estimation \sep
  educational data \sep 
  large language models \sep
  assessment
\end{keywords}

\maketitle

\section{Introduction}

Good exam design is time-consuming and difficult. One of the challenges is to ensure consistent difficulty over multiple years, as exam scores should be comparable between cohorts. As previous exams might circulate among students, instructors are required to design exams anew, selecting questions that are neither too difficult, nor too easy \cite{bachman1990fundamental}. One solution is to randomly select a sufficiently large sample of questions from an \textit{item pool} \cite{way1998protecting}.  However, the number of questions can often not be sufficiently large to be confident that the difficulty will remain constant over years. This requires instructors to estimate the difficulty of the questions to ensure consistency, a process that is often an implicit aspect of exam design. 

In this paper we assess whether Artificial Intelligence ({AI}), and in particular Natural Language Processing ({NLP}), can be used to assist instructors in this process. {AI} is being viewed as a valuable avenue for decreasing workload and increasing the capacity of educational staff in a variety of applications, ranging from tutor chat-bots to systems that can grade exams \cite{holmes2022state}.  Even though using AI for difficulty estimation has been explored \cite{benedetto2023survey}, success has been modest \cite{AlKhuzaey2024}, with {NLP} systems often performing marginally better than average-based baselines \cite{yaneva2024findings}. This task is also challenging for teachers, as shown by \citeauthor{vandvand2006b3} \cite{vandvand2006b3}. They found that teachers could correctly estimate the difficulty levels for only a small proportion of the questions.

The modest success of question difficulty estimation using NLP methods and the known limitations of teachers to estimate question difficulty motivates this study. It is clear that both teachers and NLP-based methods have a limited ability to estimate exam item difficulty, but it is not known how they compare. This comparison is critical to determining whether automated question difficulty estimation is ready for educational practice. In our work, we compare NLP-based approaches for automated question difficulty estimation with expert-human estimation of difficulty. We demonstrate that state-of-the-art NLP methods are better at question difficulty estimations than university professors, and highlight the potential of integrating NLP-based methods in the workflow of exam design.

\section{Related Work}
The task of question difficulty estimation using {NLP} methods is not new \cite{benedetto2023survey}. Already in the 1990s, traditional {AI} methods, were employed for question difficulty estimation \cite{perkins1995, boldt_1996}. More recent approaches are typically based on the transformer architecture. An example of this is the recent ``Building Educational Applications" shared task on ``Automated Prediction of Item Difficulty and Item Response Time" \cite{yaneva2024findings}, wherein a variety of approaches were explored ranging from changing the transformer architecture to data augmentation techniques. The best performing team (EduTec) used a combination of model optimisation techniques including scalar mixing \cite{gombert2023coding}, rational activation \cite{Pade2020} and multi-task learning to predict the proportion of students answering each question correctly \citep{gombert-etal-2024-predicting}. 

\subsubsection*{Our Contribution} The goal of this study is to establish whether modern {NLP}-based methods can be applied for question difficulty estimation in university education. This is operationalized by comparing whether {NLP}-based approaches perform similar or better than the lecturers who would normally construct the exams. To the best of our knowledge, this is the first study of this kind. This comparison is conducted using two university-level exams, the moderate-size question set being representative of the data that would typically be available in real-world scenarios. The code implementation of this project is publicly available.\footnote{\url{https://github.com/LeonidasZotos/nlp_vs_professors_difficulty_estimation}}

\section{Methods}
To compare the performance of professors and {LLM}-based solutions in question difficulty estimation, we collected a dataset of exam questions used in university education. The proportion of students answering a question correctly (known as the $p^+\text{-value}$) is considered the ground-truth difficulty. The professors and {LLM}-based methods estimate this ground truth based on the exam question text. We chose to use the $p^+\text{-value}$ over {IRT} metrics \cite{hambleton1991fundamentals}, because it is more intuitive for a professor to interpret and estimate.

\subsection{Exam data}
We included data from two courses in the area of Artificial Intelligence that are taught at University of Groningen. Specifically, we used Neural Networks, which is taught in the Artificial Intelligence BSc program, and Advanced Machine Learning, which is taught in the Artificial Intelligence MSc program. Both courses and exams follow a similar setup. The course material consists of custom lecture notes, and the exams are made of twenty-two questions selected from a private item pool of exam questions. Each question in the exam is a True/False question, and students have two hours to complete the exam. Examples of exam questions are shown in Table \ref{tab:exam_question_examples}.

\begin{table*}[ht]
    \caption{Two example exam questions which were given to the students as practice material.}
    \label{tab:exam_question_examples}
    \footnotesize
    \begin{tabular}{p{0.8\linewidth} l }
    \toprule
        \textbf{Exam Question} & \textbf{Answer}\\
        \midrule
        \textit{(Machine Learning basics)} Given a training data set $(\mathbf{u}_i, \mathbf{y}_i)_{i=1, ..., N}$, where $\mathbf{u}_i \in \mathbb{R}^n$ and $\mathbf{y}_i \in \mathbb{R}^m$, then for any model $f: \mathbb{R}^n \to \mathbb{R}^m$ and any loss function $L$, the empirical risk $\mathcal{R}^{emp}(f)$ is less or equal to the risk $\mathcal{R}(f)$.    & False \\\midrule
        \textit{(Elementary math)} Let $f, g: \mathbb{R}^n \to \mathbb{R}$ be differentiable functions with gradients $\nabla f, \nabla g$. Then $\nabla(f+g) = \nabla f + \nabla g$. & True \\
    \bottomrule
    \end{tabular}
\end{table*}

We collected three archived exams for each course, covering the years 22/23 ($111$ students), 23/24 ($114$ students), and 24/25 ($20$ students) for Neural Networks and the years 21/22 ($103$ students), 22/23 ($119$ students) and 23/24 ($71$ students) for Advanced Machine Learning. We collected all questions from the exams and pooled them together. For questions that were repeated across years, the $p^+\text{-value}$ was based on all students that received this question. This was the case for five questions in total for each of the two courses. Moreover, the examiner of the courses considered four questions from Advanced Machine Learning and one question from Neural Networks as ambiguous and marked those as correct for both True and False. Those ambiguous questions were removed from this study. Additionally, there was one question which included an image. This question was also removed as we consider this to be out-of-scope for the present study. 
This resulted in $59$ questions from Neural Networks and $53$ questions from Advanced Machine Learning.

We use this new dataset instead of using existing datasets for three reasons. First and foremost, in contrast to other datasets, we have professors that are experts in the field available that can provide $p^+\text{-value}$ estimates to represent the manual question difficulty estimation in a way that is ecologically valid. Secondly, the questions in our new dataset are not publicly available guaranteeing that they are unseen for all {LLMs}. Finally, by analyzing how difficulty estimation methods perform on questions involving abstract mathematical reasoning and comprehension, we examine whether their previous success in assessing the difficulty of clinical decision-making and language comprehension exams \cite{yaneva2024findings, cmcqrd_2023} extends to this domain.

\subsection{Professors' estimations}
\label{sec:professor_annotations}
Four professors of the University of Groningen were asked to estimate for each question the percentage of students that would answer correctly. All four professors have expertise in Machine Learning and Neural Networks and would be qualified to teach these courses. One of the four professors (henceforth referred to as Professor 0) was the person who created the exams, has taught the courses and is familiar with the student cohorts that took the exams. The remaining professors have not taught these specific courses, and have never been a student in these courses. This ensures they are fairly knowledgeable about the population of students and the topic, but have not seen students' performance on these questions, in contrast to Professor 0. As an example and to provide some calibration, the professors were given one exam question with the true percentage of students that answered it correctly. The professors were also given the correct True/False answers. This was to help them focus on the task --- predicting the difficulty of the question rather than solving it. The exact annotation instructions are presented in Table \ref{tab:annotation_instruction}.

\begin{table*}[ht]
    \caption{Instructions given to the professors. Equivalent instructions were given for the Neural Networks exam.}
    \label{tab:annotation_instruction}
    \footnotesize
    \begin{tabular}{p{0.8\linewidth}}
        \toprule
        \textbf{Annotation Instruction: } Below are exam questions from the Advanced Machine Learning Course, taught in the University of Groningen. For each question, the correct answer is highlighted in \textcolor{ForestGreen}{\textbf{green}}. Estimate, from the examiner’s perspective, what percentage of students will answer each question correctly. Feel free to re-visit and adjust previous estimates. An example is presented below, where the percentage of students who selected the correct answer is provided. At the end, provide an estimate of the time spent on this item difficulty estimation task.

        \textbf{Purpose of the Study:} This study aims to compare the performance of expert educators and state-of-the-art {LLM}-based methods in estimating the difficulty of True/False questions.\\
        \bottomrule
    \end{tabular}
\end{table*}

Each professor made their estimates independently and at a moment that fits their schedule. On average, estimating the difficulty of the total of $112$ questions took each professor $2$ hours and $15$ minutes. One professor (professor $3$) declined to give estimates for sixteen questions for Neural Networks and six questions for Advanced Machine Learning, stating that they \textit{miss the specific knowledge of some concepts to provide a confident estimate}. These questions were not considered in the evaluation for this professor, but were maintained for the rest of the analysis. 

\subsection{{NLP} Approaches}
We focus on two types of {NLP}-based methods for item difficulty estimation. We investigate methods based on prompting where {LLMs} directly estimate the question difficulty and methods based on the uncertainty of an {LLM} attempting to solve the question. The mathematical notation in all questions is encoded using LaTeX, which {LLMs} can process well \cite{frieder2023mathematical, ahn2024largelanguagemodelsmathematical}.

\paragraph{Using Direct Estimation}
As a simple comparison between {LLMs} and professors, we tested two setups in which a powerful {LLM} is prompted to directly estimate the $p^+\text{-value}$ of a question. We provided the LLMs with the same instruction (and example question) that we also gave to the professors. Additionally, we use Chain of Thought (by instructing the {LLM} to ``Think step by step") to allow the model to ``reason" before giving an estimate \cite{wei2023chainofthoughtpromptingelicitsreasoning}. To get the most competitive results we use \texttt{gemini-2.5-pro-preview-03-25} (\texttt{Gemini 2.5}) and \texttt{Gemini-2.0-flash} (\texttt{Gemini 2.0}), two of the best-performing current {LLMs}, as measured by the community-driven Chatbot Arena \cite{chiang2024chatbot}. At the time of writing, they rank 1st and 8th respectively. 

The {LLMs} are prompted using two different setups. In the single question setup, the {LLM} is tasked with predicting the $p^+\text{-value}$ of each question individually without being able to see the other questions. In contrast, in the all-questions setup the LLM is given the complete question set and is tasked with estimating the $p^+\text{-values}$ of all items in one go. We consider both of these setups to be promising, each with its own trade-offs. On one hand, we observe that prompting the model to generate the difficulty of a single question item encourages it to generate longer reasoning streams, which could lead to more accurate predictions. At the same time, predicting the difficulty of all items concurrently can also be beneficial, potentially steering the prediction of each item to be informed by the entire set. The all-questions setup closely resembles the setup with the professors, as they can also see all questions to gauge the overall difficulty of the question set. 

\paragraph{Using {LLM} Uncertainty}
\label{sec:using_LLM_uncertainty}
As a task-specific question difficulty estimation method we implement the approach by \citeauthor{zotos2024doubtful} \cite{zotos2024doubtful}, which is a good representation of the current state-of-the-art for this task.

In this approach, a set of nine {LLMs} are prompted to solve each question without the answer, where the uncertainty of the {LLM} can be used as a feature for a supervised learning model to predict the $p^+\text{-value}$. By using a mixture of stronger and weaker {LLMs} we can get a good spread of {LLM} uncertainties as features. We use the same {LLMs} as in the original work.

Two measures of uncertainty are used to indicate the difficulty of the question according to the {LLM}. One is the  probability of the first generated token (probability of ``A" or ``B"). The other one is Choice-Order Sensitivity \cite{pezeshkpour2023large}, which measures whether the {LLM} gives the same prediction when the order of the answer choices is shuffled. This is operationalised by performing inference with different permutations of choice-order (for True/False questions, only two permutations are possible) and calculating the proportion of times the correct answer is selected. Both measures have been found to correlate with the probability that a prediction from an {LLM} is correct \cite{plaut2024softmax, zotos2024doubtful} as well as the $p^+\text{-values}$ of exam questions \cite{zotos-etal-2025-model}.

As supervised learning models we use three different regressors: Random Forest, Support Vector Machine and Linear Regression. Each model is trained for each course using an $80$:$20$ train-test split. The regression models for Neural Networks are therefore trained with $47$ samples, and the models for Advanced Machine Learning with $42$ samples. Using Grid Search with $5$-fold validation we determine the best hyperparameters for each regression model. 

We compare this with two alternative supervised learning setups. In one we train a dummy model with no features, always predicting the mean $p^+\text{-value}$. In the other, we consider what may be learned with simple features from the text. For this we use {TF-IDF} features for a supervised learning model. For completeness, we also consider the concatenation of {TF-IDF} features and {LLM} uncertainties as features. 

Arguably, comparing the professors to this supervised learning approach is unfair. The professors are only given one ``labeled" example to calibrate their predictions, while the regressor needs more than one example to undergo training. The supervised-learning approach therefore can directly learn the distribution of $p^+\text{-values}$, which the professors do not have access to. 
At the same time, this setup is realistic for an advanced {NLP} setup that may be used in practice. Universities often have archived data of previous years' exams, but reviewing them is time-consuming. Using this supervised learning setup, we can capitalize on this existing data effectively.

\section{Results}

To evaluate how the professors compare to the {NLP}-based methods on the question difficulty estimation, we use the root mean squared error ({RMSE}) between the estimated $p^+\text{-values}$ and the ground truth values as a standard metric for error. We also measure the rank correlation between the estimates and the ground truth using Spearman’s $\rho$. This rank correlation assessment allows us to detect whether an approach which has consistently biased estimates still maintains a strong monotonic relationship with the true $p^+\text{-values}$ and is thus able to distinguish easy from difficult questions. The Mean Error ({ME}) metric directly evaluates any consistent bias, by measuring whether the difficulty estimates are on average too high or too low. The results of all experiments are presented in Table \ref{tab:results:main_results}. 

\begin{table*}[t]
    \caption{Root Mean Squared error ({RMSE}), Mean Error ({ME}) and Spearman's $\rho$ as measured for each Item Difficulty Estimation Approach. Overall best performance is indicated with \textbf{boldface}, while \textit{italics} indicates best in category.}
    \label{tab:results:main_results}
    \adjustbox{max width=\textwidth}{
    \footnotesize
    \begin{tabular}{llrrrrrr}
        \toprule
        \multirow{2}{*}{\textbf{Method}} & \multirow{2}{*}{\textbf{Features}} & \multicolumn{3}{c}{\textbf{NN}}   & \multicolumn{3}{c}{\textbf{AML}}      \\  
                          \cmidrule(lr){3-5} \cmidrule(lr){6-8}
                         & & RMSE $\downarrow$ & ME     & $\rho$ $\uparrow$    & RMSE $\downarrow$ & ME    & $\rho$ $\uparrow$     \\
        \midrule
        \hlgray{\textbf{Professors}}                                                    \\
        Professor 0 (Exam Creator) &  & 0.192 & 0.051	    & -0.041                                                                                       & 0.160 & 0.038	    & 0.392                     \\ \hdashline
        Professor 1     &  & 0.336 & 0.048	    & -0.014                                                                                       & 0.357 & -0.087	    & 0.062                     \\
        Professor 2     &  & 0.245 & -0.038	    & -0.011                                                                                       & \textit{0.185} & 0.070	    & \textit{0.241}                     \\
        Professor 3     &  & \textit{0.184} & 0.000 & \textit{0.211}  & 0.193 & -0.003 & 0.224                                          \\
        Average of Professors' Estimates &  & 0.234  & 0.001	    & -0.020                                                                                          & 0.205 & -0.010 & 0.173                                       \\
        
        \midrule
        \hlgray{\textbf{LLM Direct Estimation}}  \\
        \multirow{2}{*}{\texttt{Gemini 2.0}} 
         & Single question & 0.242 & -0.071	& 0.118        & 0.199 & -0.061	& 0.188 \\
         & All questions & 0.218 & 0.016	& 0.062                                     & 0.307 & -0.177	& 0.052 \\ 
         \multirow{2}{*}{\texttt{Gemini 2.5}} 
         & Single question & \textit{0.192} & 0.035  & \textit{0.283}  & \textit{0.165} & 0.044  & \textit{0.345} \\
         & All questions & 0.199 & 0.034  	& 0.175         & 0.243	& 0.051  & 0.079 \\
        \midrule
         
        \hlgray{\textbf{Supervised Learning}}                                         \\
        Dummy Model  & None & 0.174 & -   & -                     & 0.126 & -    & -                         \\
        \hdashline
        \multirow{3}{*}{Linear Regression} 
        & LLM Uncertainties         & 0.165 & 0.024 &  0.678                    & 0.119 &  0.004 & 0.418 \\        
        & TF-IDF                    & 0.174 & 0.016 &  0.098                    & 0.126 & -0.000 & 0.251 \\
        & TF-IDF \& Uncertainties   & 0.165 & 0.024 &  0.741                    & 0.119 &  0.003 & 0.464 \\
        \hdashline
        \multirow{3}{*}{Random Forest} 
        & LLM Uncertainties         & 0.170 & 0.032 & 0.273                     & 0.120 & -0.009 & 0.282 \\
        & TF-IDF                    & 0.180 & 0.004 & 0.035                     & 0.115 & 0.003 & 0.469 \\
        & TF-IDF \& Uncertainties   & 0.179 &  0.011 & 0.049                    & 0.109 & 0.004 & 0.464\\
        \hdashline
        \multirow{3}{*}{Support Vector Machine}
        & LLM Uncertainties         & 0.148 & 0.011 & 0.811                     & \textbf{0.107} & 0.009 & \textbf{0.582} \\
        & TF-IDF                    & 0.174 & 0.018 & -0.070                    & 0.117 & 0.008 & 0.314 \\
        & TF-IDF \& Uncertainties   & \textbf{0.147} & 0.001 & \textbf{0.853}   & 0.109 & 0.013 & 0.491 \\
        \bottomrule
    \end{tabular}}
\end{table*}

\paragraph{Professor Performance}
 Overall, and in line with the study by \citeauthor{vandvand2006b3} \cite{vandvand2006b3}, professors seem to have limited ability to estimate question difficulty. We see that for the Neural Networks ({NN}) exam, three of the professors (including the exam creator) estimated $p^+\text{-values}$ that do not correlate with the performance of students. Only professor $3$ has a positive rank correlation with $\rho{=}0.211$. This may be partly because professor $3$ did not give an answer to sixteen questions for Neural Networks, presumably the ones they felt uncertain about. For the MSc level Advanced Machine Learning ({AML}) course professor $2$ and professor $3$ achieved better performances, but both underperformed compared to Professor 0, whose estimates moderately correlate with the true  $p^+\text{-values}$.

 The Mean Errors are sometimes positive and sometimes negative, depending on the professor and the exam. This suggests that there is no clear pattern of professors consistently over/underestimating question difficulty. The high {RMSE} does show that overall the professors perform poorly at directly estimating $p^+\text{-values}$. Furthermore, for each question we also average the three professor estimates (excluding Professor 0) as if they are voting. This did not lead to any improvements.

 \paragraph{Direct Prompting Performance}
  When assessing the two methods of direct prompting, we find that prompting the model with one question at a time generally leads to lower {RMSE} and higher rank correlations, with the exception of \texttt{Gemini 2.0} in the Neural Networks set. Additionally, we find that \texttt{Gemini 2.5} is consistently more accurate than \texttt{Gemini 2.0} which corresponds well with their performance in other tasks \cite{chiang2024chatbot}. When comparing the direct prompting of LLMs to the professors we find that the LLMs tend to perform better. The best {LLM} method (\texttt{Gemini 2.5}, single question) has better rank correlation than all professors on both exams, except when compared to Professor 0, who performs better in the Advanced Machine Learning set.

\paragraph{Supervised Learning Performance} 
 The supervised learning methods achieve lower {RMSE} than the professors and the direct {LLM} predictions, because only the supervised learning methods are able to learn the distribution of $p^+\text{-values}$ from data. The SVM performs best likely due to the small dataset and non-linear relationships. Using only {TF-IDF} features was not sufficient to estimate $p^+\text{-values}$ for the Neural Networks set, but was already better than the professors and often better than the {LLMs} for the Advanced Machine Learning set. The {LLM} uncertainties as features are substantially more predictive, resulting in lower RMSE and higher rank correlation $\rho$. The {SVM} with {TF-IDF} Scores and {LLM} Uncertainties performed the best, with a rank correlation of $\rho{=}0.853$ for Neural Networks. For Advanced Machine Learning, the {SVM} trained only on {LLM} Uncertainties performed best, with a rank correlation of $\rho{=}0.582$. This is much better than either direct estimation from the {LLM}, or estimation from the professors. 
  
\subsection{Inter-Annotator Agreement}
\begin{figure*}[ht!]
    \centering
    \begin{subfigure}[b]{0.4\textwidth}
        \centering
        \includegraphics[width=\linewidth]{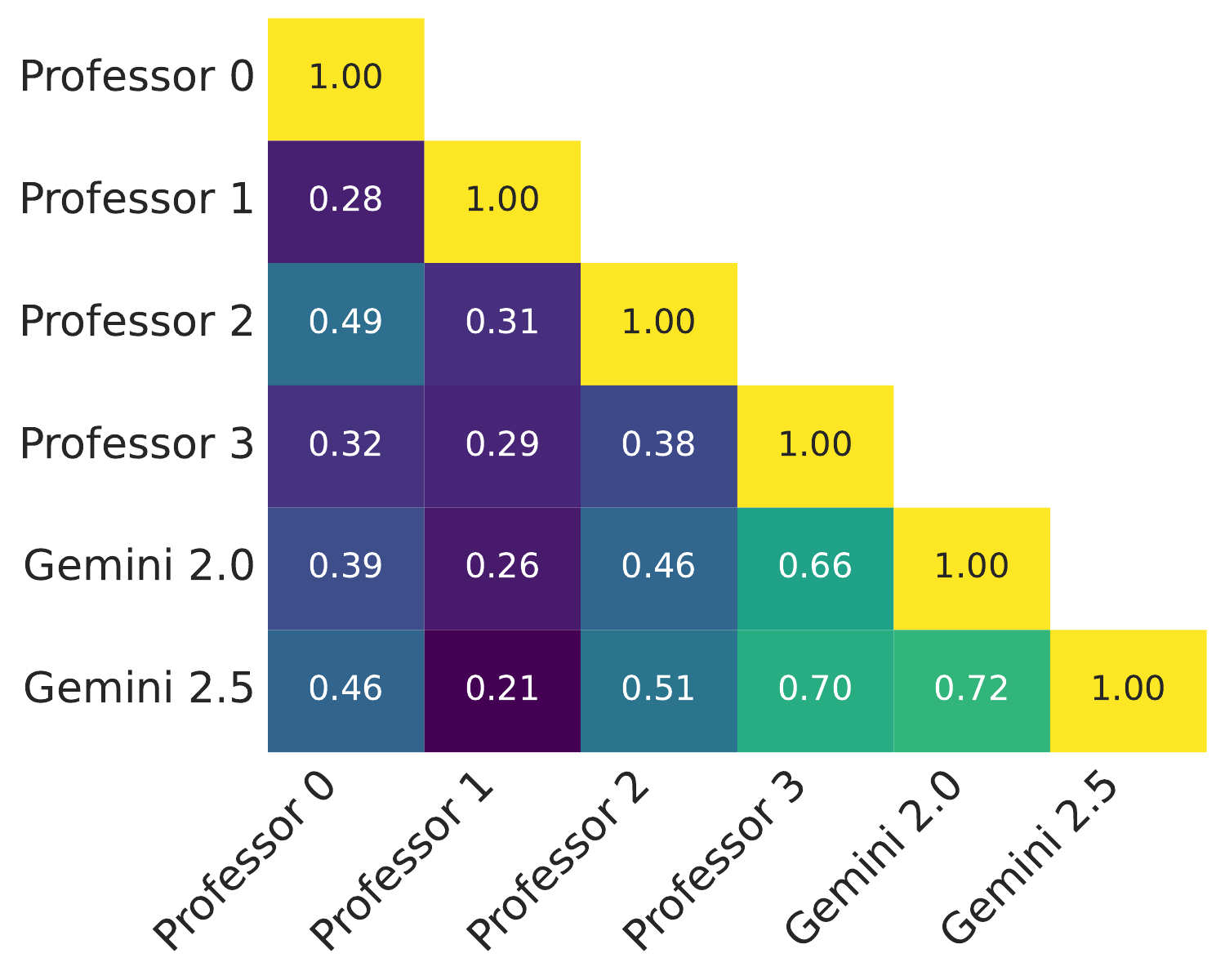}%
        \caption{Neural Networks}
        \label{subfig:nn_correlation}
    \end{subfigure}%
    \hspace{0.05\textwidth}
    \begin{subfigure}[b]{0.4\textwidth}
        \centering
        \includegraphics[width=\linewidth]{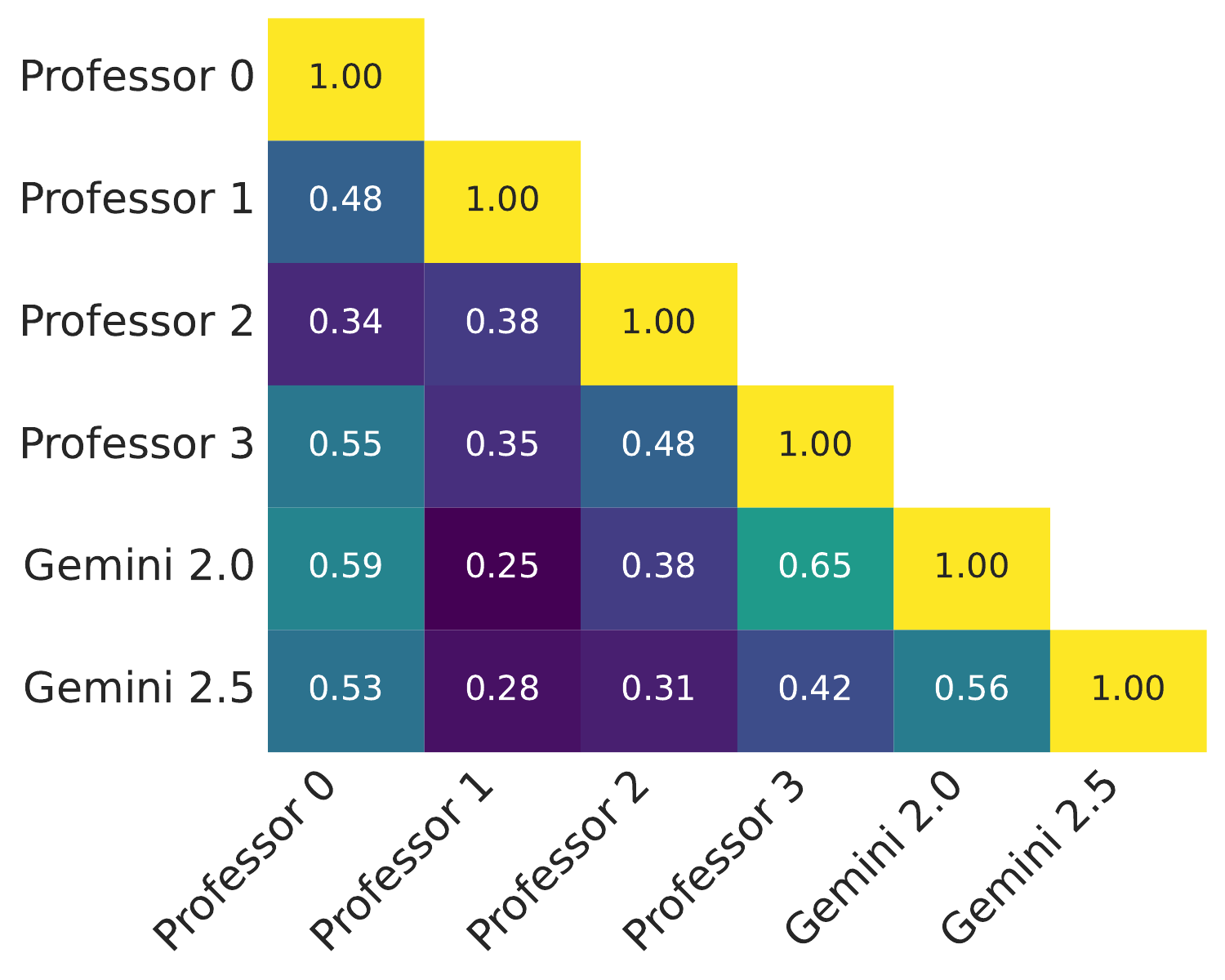} %
        \caption{Advanced Machine Learning}
        \label{subfig:aml_correlation}
    \end{subfigure}
    \caption{Inter-Annotator Agreement, calculated using Spearman's $\rho$ for the two datasets.}
    \label{fig:assessment_correlations}
\end{figure*}

Figure \ref{fig:assessment_correlations} shows the inter-annotator agreement (including the direct assessment of the Gemini {LLMs}), represented using the Spearman correlation coefficient. Overall, this analysis shows that, while the task is difficult (as was shown earlier), there are moderate correlations between the professors, indicating that they might be over/under-estimating the difficulty of the same questions. This suggests that for Advanced Machine Learning professors have a consistent notion of what should be difficult and what should be easy and are making informed estimates.

Additionally, we observe a high correlation between professor $3$ and the Gemini Models in the Neural networks dataset, in line with their relatively good performance on the set. Lastly, we also find a moderate to high correlation between the assessments of the two Gemini {LLMs}, suggesting that {LLMs} of the same family behave consistently on this task.

\subsection{Per Question Analysis}
\label{sec:per_question_analysis}
\begin{figure*}[t!]
    \centering 
    \begin{subfigure}[b]{0.86\textwidth} 
        \centering
        \includegraphics[width=\linewidth]{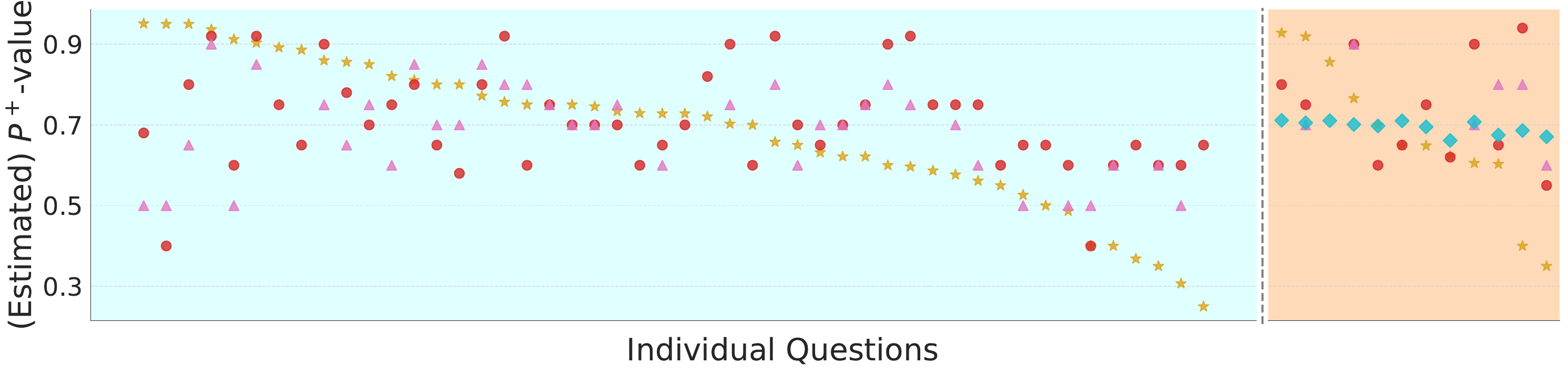} 
        \caption{Neural Networks}
        \label{subfig:nn_estimates}
    \end{subfigure}
    \vspace{\floatsep} 
    \begin{subfigure}[b]{\textwidth}
        \centering
        \includegraphics[width=0.86\linewidth]{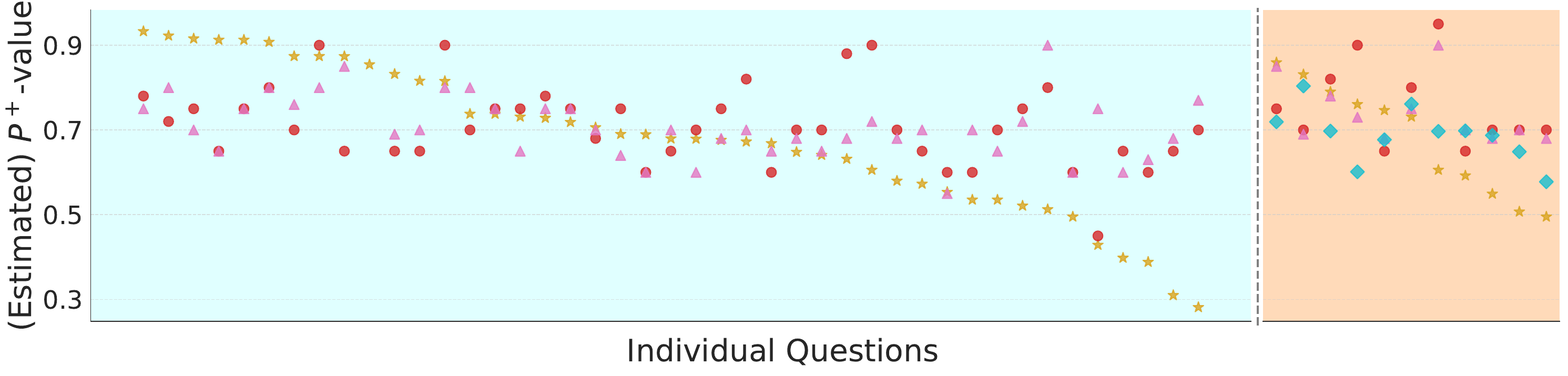}
        \caption{Advanced Machine Learning}
        \label{subfig:aml_estimates}
    \end{subfigure}

    \vspace{1ex} 
    \includegraphics[width=0.65\linewidth]{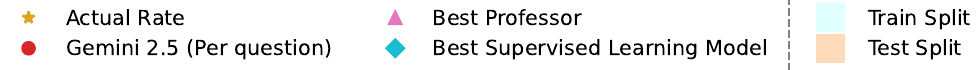} 

    \caption{Estimated $p^+\text{-value}$ per question item for the best-performing professor, \texttt{Gemini 2.5} and trained support vector machine using {LLM} Uncertainties. The Train/Test split is only relevant for the Supervised Learning Model, for which we only report its performance on the unseen question items.}
    \label{fig:estimates_overview}
\end{figure*}

Figure \ref{fig:estimates_overview} presents, per question, the $p^+\text{-value}$, along with the estimates of the best performing systems per category. For each dataset, we separate the questions based on the train and test splits used for the best-performing Supervised Learning approach (this separation has no impact on the teachers and prompted {LLMs}). Here, we directly observe that there is a good range and distribution of difficulties, with a balance of easy and difficult questions. We also see that a few questions in each set were answered correctly by less than $40$\% of the student population, suggesting that these questions might be misleading or trick-questions. 

Looking at the general picture, and in line with the results presented in Table \ref{tab:results:main_results}, the best professors' predictions and \texttt{Gemini 2.5}'s predictions do not show a strong correlation with the true $p^+\text{-values}$. Additionally, their estimates show high variability, but are seldom below $50$\%, suggesting that they do not recognize trick-questions that might lead students to perform worse than random guessing. In contrast, the estimated $p^+\text{-values}$ of the best trained Supervised Learning Model show low variability and remain near the average $p^+\text{-value}$ observed in the training sets. We do see that the estimated $p^+\text{-values}$ correlate with the true $p^+\text{-values}$, but that the very easy/difficult questions are estimated close to the average. This explains the good rank correlation, but still high RMSE that we observed in Table \ref{tab:results:main_results}.

\section{Discussion}
\label{sec:discussion}
To conclude, in our study we show that strong LLMs estimated exam difficulty at least as well as professors that were not directly involved in the course, while the trained regressor estimated exam difficulty better than the professor who created the exams.
 
Specifically, we have shown that \texttt{Gemini 2.5} is better at question difficulty estimation for Neural Networks and Advanced Machine Learning exams than three university professors. We also find that \texttt{Gemini 2.5} consistently outperforms \texttt{Gemini 2.0} on this task, which suggests that future {LLM} releases may lead to further improvement. 

Additionally, we find that with as little as $42$ training samples the supervised learning method of \citeauthor{zotos-etal-2025-model} \cite{zotos-etal-2025-model} based on the uncertainty of {LLMs} solving the problem  substantially outperforms both professors and a standard {LLM} approach. This finding is significant, as it demonstrates that implementing this  system for individual courses is feasible, with only a couple of exams from previous years being required to train a good regression model.

Lastly, our findings are on questions that require parsing mathematical notation and that require mathematical reasoning. This extends previous successes of {NLP}-based question difficulty estimation on biopsychology \cite{zotos-etal-2025-model}, clinical decision making \cite{yaneva2024findings} and language comprehension \cite{cmcqrd_2023} exams to more mathematical fields. While the current results are specific to Machine Learning, they suggest that {NLP} methods are also promising to other mathematical topics such as physics, computer science and astronomy. 

Overall we have demonstrated that state-of-the-art {NLP} methods are -- relative to professors -- very good at question difficulty estimation, and can support them in ranking question difficulty. Of course our findings are focused on question difficulty estimation, and we still need professors for the many other aspects of exam design and education! 

\paragraph{Limitations}
The primary limitation of this study is that three of the four professors were not involved in the courses. Naturally, the professor who was actively involved in teaching the courses has additional information which can help with the task of question difficulty estimation, such as knowledge of which subjects are more extensively covered in the lectures. At the same time, all professors in our study already have significantly more background information than the better-performing \texttt{Gemini 2.5}, as they are familiar with the whole curriculum and know how these students perform in other courses. Due to the small scale of this study, its findings cannot be generalized to other educational fields or different types of assessments. Therefore, we want to emphasize that while NLP methods surpassed the performance of some professors in this specific task, it should not be concluded that NLP outperforms all professors in this area.

We also observed that professors mostly make $p^+\text{-value}$ predictions in increments of 5\% (e.g., 65\% or 70\%, but not 68\%). This results in items being tied in terms of predicted $p^+\text{-value}$. While these ties do not impact the calculation of the Root Mean Squared Error, they might negatively affect the Spearman Rank Correlation Coefficient, as granular judgments that would create a clear ranking of the question items is not available. However, we observe that \texttt{Gemini 2.5}'s direct estimations also frequently occur in 5\% increments (with the model often predicting 60\% or 75\%, as shown in Figure \ref{fig:estimates_overview}), yet a consistently higher correlation is observed compared to the professors' annotations.

\paragraph{Broader Impact Statement}
While the results of the current study are promising, implementing such a system is not trivial: The best performing system we tested relies on the existence of some training data, as well as the availability of sufficient computational resources to compute the uncertainty metrics of the {LLMs}. At the same time, instructing a state-of-the-art proprietary {LLM} to estimate question difficulty can lead to good performance on the task, a solution that is trivial to use. As a final consideration, we believe that any system of this type should be used in a human-in-the-loop fashion to address cases where the {NLP} methods unavoidably lack context such as, for example, when a question is assessed as easy even though the material was not covered in class.

\begin{acknowledgments}
We would like to thank Professor Herbert Jaeger for providing the exam material for this study. We would also like to thank the three professors for volunteering their time and expertise to provide estimates of question difficulty. This study was positively reviewed by the Faculty of Science and Engineering Ethics Committee under reference {FSE.EC25005}.
\end{acknowledgments}

\section*{Declaration on Generative AI}
For the preparation of this work, Gemini 2.5 was used for: Grammar check.

\bibliography{sample-ceur}

\begin{thebibliography}{22}
\expandafter\ifx\csname natexlab\endcsname\relax\def\natexlab#1{#1}\fi
\providecommand{\url}[1]{\texttt{#1}}
\providecommand{\href}[2]{#2}
\providecommand{\path}[1]{#1}
\providecommand{\DOIprefix}{doi:}
\providecommand{\ArXivprefix}{arXiv:}
\providecommand{\URLprefix}{URL: }
\providecommand{\Pubmedprefix}{pmid:}
\providecommand{\doi}[1]{\href{http://dx.doi.org/#1}{\path{#1}}}
\providecommand{\Pubmed}[1]{\href{pmid:#1}{\path{#1}}}
\providecommand{\bibinfo}[2]{#2}
\ifx\xfnm\relax \def\xfnm[#1]{\unskip,\space#1}\fi
\bibitem[{Bachman(1990)}]{bachman1990fundamental}
\bibinfo{author}{L.~F. Bachman}, \bibinfo{title}{Fundamental considerations in language testing}, \bibinfo{publisher}{Oxford university press}, \bibinfo{year}{1990}.
\bibitem[{Way(1998)}]{way1998protecting}
\bibinfo{author}{W.~D. Way},
\newblock \bibinfo{title}{Protecting the integrity of computerized testing item pools},
\newblock \bibinfo{journal}{Educational Measurement: Issues and Practice} \bibinfo{volume}{17} (\bibinfo{year}{1998}) \bibinfo{pages}{17--27}.
\bibitem[{Holmes and Tuomi(2022)}]{holmes2022state}
\bibinfo{author}{W.~Holmes}, \bibinfo{author}{I.~Tuomi},
\newblock \bibinfo{title}{State of the art and practice in {AI} in education},
\newblock \bibinfo{journal}{European journal of education} \bibinfo{volume}{57} (\bibinfo{year}{2022}) \bibinfo{pages}{542--570}.
\bibitem[{Benedetto et~al.(2023)Benedetto, Cremonesi, Caines, Buttery, Cappelli, Giussani, and Turrin}]{benedetto2023survey}
\bibinfo{author}{L.~Benedetto}, \bibinfo{author}{P.~Cremonesi}, \bibinfo{author}{A.~Caines}, \bibinfo{author}{P.~Buttery}, \bibinfo{author}{A.~Cappelli}, \bibinfo{author}{A.~Giussani}, \bibinfo{author}{R.~Turrin},
\newblock \bibinfo{title}{A survey on recent approaches to question difficulty estimation from text},
\newblock \bibinfo{journal}{ACM Computing Surveys} \bibinfo{volume}{55} (\bibinfo{year}{2023}) \bibinfo{pages}{1--37}.
\bibitem[{AlKhuzaey et~al.(2024)AlKhuzaey, Grasso, Payne, and Tamma}]{AlKhuzaey2024}
\bibinfo{author}{S.~AlKhuzaey}, \bibinfo{author}{F.~Grasso}, \bibinfo{author}{T.~R. Payne}, \bibinfo{author}{V.~Tamma},
\newblock \bibinfo{title}{Text-based question difficulty prediction: {A} systematic review of automatic approaches},
\newblock \bibinfo{journal}{International Journal of Artificial Intelligence in Education} \bibinfo{volume}{34} (\bibinfo{year}{2024}) \bibinfo{pages}{862--914}. \DOIprefix\doi{10.1007/s40593-023-00362-1}.
\bibitem[{Yaneva et~al.(2024)Yaneva, North, Baldwin, Ha, Rezayi, Zhou, Choudhury, Harik, and Clauser}]{yaneva2024findings}
\bibinfo{author}{V.~Yaneva}, \bibinfo{author}{K.~North}, \bibinfo{author}{P.~Baldwin}, \bibinfo{author}{L.~A. Ha}, \bibinfo{author}{S.~Rezayi}, \bibinfo{author}{Y.~Zhou}, \bibinfo{author}{S.~R. Choudhury}, \bibinfo{author}{P.~Harik}, \bibinfo{author}{B.~Clauser},
\newblock \bibinfo{title}{Findings from the first shared task on automated prediction of difficulty and response time for multiple-choice questions},
\newblock in: \bibinfo{booktitle}{Proceedings of the 19th Workshop on Innovative Use of {NLP} for Building Educational Applications ({BEA} 2024)}, \bibinfo{year}{2024}, pp. \bibinfo{pages}{470--482}.
\bibitem[{van~de Watering and van~der Rijt(2006)}]{vandvand2006b3}
\bibinfo{author}{G.~van~de Watering}, \bibinfo{author}{J.~van~der Rijt},
\newblock \bibinfo{title}{Teachers’ and students’ perceptions of assessments: {A} review and a study into the ability and accuracy of estimating the difficulty levels of assessment items},
\newblock \bibinfo{journal}{Educational Research Review} \bibinfo{volume}{1} (\bibinfo{year}{2006}) \bibinfo{pages}{133--147}. \URLprefix \url{https://www.learntechlib.org/p/197391}.
\bibitem[{Perkins et~al.(1995)Perkins, Gupta, and Tammana}]{perkins1995}
\bibinfo{author}{K.~Perkins}, \bibinfo{author}{L.~Gupta}, \bibinfo{author}{R.~Tammana},
\newblock \bibinfo{title}{Predicting item difficulty in a reading comprehension test with an artificial neural network},
\newblock \bibinfo{journal}{Language Testing} \bibinfo{volume}{12} (\bibinfo{year}{1995}) \bibinfo{pages}{34--53}. \DOIprefix\doi{10.1177/026553229501200103}.
\bibitem[{Boldt and Freedle(1996)}]{boldt_1996}
\bibinfo{author}{R.~F. Boldt}, \bibinfo{author}{R.~Freedle},
\newblock \bibinfo{title}{Using a neural net to predict item difficulty},
\newblock \bibinfo{journal}{ETS Research Report Series}  (\bibinfo{year}{1996}) \bibinfo{pages}{i--19}. \DOIprefix\doi{https://doi.org/10.1002/j.2333-8504.1996.tb01709.x}.
\bibitem[{Gombert et~al.(2023)Gombert, Mitri, Karademir, Kubsch, Kolbe, Tautz, Grimm, Bohm, Neumann, and Drachsler}]{gombert2023coding}
\bibinfo{author}{S.~Gombert}, \bibinfo{author}{D.~D. Mitri}, \bibinfo{author}{O.~Karademir}, \bibinfo{author}{M.~Kubsch}, \bibinfo{author}{H.~Kolbe}, \bibinfo{author}{S.~Tautz}, \bibinfo{author}{A.~Grimm}, \bibinfo{author}{I.~Bohm}, \bibinfo{author}{K.~Neumann}, \bibinfo{author}{H.~Drachsler},
\newblock \bibinfo{title}{Coding energy knowledge in constructed responses with explainable {NLP} models},
\newblock \bibinfo{journal}{Journal of Computer Assisted Learning} \bibinfo{volume}{39} (\bibinfo{year}{2023}) \bibinfo{pages}{767--786}.
\bibitem[{Molina et~al.(2020)Molina, Schramowski, and Kersting}]{Pade2020}
\bibinfo{author}{A.~Molina}, \bibinfo{author}{P.~Schramowski}, \bibinfo{author}{K.~Kersting},
\newblock \bibinfo{title}{Pad{\'{e}} activation units: End-to-end learning of flexible activation functions in deep networks},
\newblock in: \bibinfo{booktitle}{8th International Conference on Learning Representations, {ICLR}, Addis Ababa, Ethiopia, April 26-30, 2020}, \bibinfo{publisher}{OpenReview.net}, \bibinfo{year}{2020}. \URLprefix \url{https://openreview.net/forum?id=BJlBSkHtDS}.
\bibitem[{Gombert et~al.(2024)Gombert, Menzel, Mitri, and Drachsler}]{gombert-etal-2024-predicting}
\bibinfo{author}{S.~Gombert}, \bibinfo{author}{L.~Menzel}, \bibinfo{author}{D.~D. Mitri}, \bibinfo{author}{H.~Drachsler},
\newblock \bibinfo{title}{Predicting item difficulty and item response time with scalar-mixed transformer encoder models and rational network regression heads},
\newblock in: \bibinfo{editor}{E.~Kochmar}, \bibinfo{editor}{M.~Bexte}, \bibinfo{editor}{J.~Burstein}, \bibinfo{editor}{A.~Horbach}, \bibinfo{editor}{R.~Laarmann-Quante}, \bibinfo{editor}{A.~Tack}, \bibinfo{editor}{V.~Yaneva}, \bibinfo{editor}{Z.~Yuan} (Eds.), \bibinfo{booktitle}{Proceedings of the 19th Workshop on Innovative Use of {NLP} for Building Educational Applications ({BEA} 2024)}, \bibinfo{publisher}{Association for Computational Linguistics}, \bibinfo{address}{Mexico City, Mexico}, \bibinfo{year}{2024}, pp. \bibinfo{pages}{483--492}. \URLprefix \url{https://aclanthology.org/2024.bea-1.40}.
\bibitem[{Hambleton et~al.(1991)Hambleton, Swaminathan, and Rogers}]{hambleton1991fundamentals}
\bibinfo{author}{R.~K. Hambleton}, \bibinfo{author}{H.~Swaminathan}, \bibinfo{author}{H.~J. Rogers}, \bibinfo{title}{Fundamentals of item response theory}, volume~\bibinfo{volume}{2}, \bibinfo{publisher}{Sage}, \bibinfo{year}{1991}.
\bibitem[{Mullooly et~al.(2023)Mullooly, Andersen, Benedetto, Buttery, Caines, Gales, Karatay, Knill, Liusie, Raina, and Taslimipoor}]{cmcqrd_2023}
\bibinfo{author}{A.~Mullooly}, \bibinfo{author}{{\O}.~Andersen}, \bibinfo{author}{L.~Benedetto}, \bibinfo{author}{P.~Buttery}, \bibinfo{author}{A.~Caines}, \bibinfo{author}{M.~J.~F. Gales}, \bibinfo{author}{Y.~Karatay}, \bibinfo{author}{K.~Knill}, \bibinfo{author}{A.~Liusie}, \bibinfo{author}{V.~Raina}, \bibinfo{author}{S.~Taslimipoor}, \bibinfo{title}{The {C}ambridge Multiple-Choice Questions Reading Dataset}, \bibinfo{publisher}{Cambridge University Press and Assessment}, \bibinfo{year}{2023}. \DOIprefix\doi{10.17863/CAM.102185}.
\bibitem[{Frieder et~al.(2023)Frieder, Pinchetti, Chevalier, Griffiths, Salvatori, Lukasiewicz, Petersen, and Berner}]{frieder2023mathematical}
\bibinfo{author}{S.~Frieder}, \bibinfo{author}{L.~Pinchetti}, \bibinfo{author}{C.~Chevalier}, \bibinfo{author}{R.-R. Griffiths}, \bibinfo{author}{T.~Salvatori}, \bibinfo{author}{T.~Lukasiewicz}, \bibinfo{author}{P.~Petersen}, \bibinfo{author}{J.~Berner},
\newblock \bibinfo{title}{Mathematical capabilities of {ChatGPT}},
\newblock in: \bibinfo{editor}{A.~Oh}, \bibinfo{editor}{T.~Naumann}, \bibinfo{editor}{A.~Globerson}, \bibinfo{editor}{K.~Saenko}, \bibinfo{editor}{M.~Hardt}, \bibinfo{editor}{S.~Levine} (Eds.), \bibinfo{booktitle}{Advances in Neural Information Processing Systems}, volume~\bibinfo{volume}{36}, \bibinfo{publisher}{Curran Associates, Inc.}, \bibinfo{year}{2023}, pp. \bibinfo{pages}{27699--27744}. \URLprefix \url{https://proceedings.neurips.cc/paper_files/paper/2023/file/58168e8a92994655d6da3939e7cc0918-Paper-Datasets_and_Benchmarks.pdf}.
\bibitem[{Ahn et~al.(2024)Ahn, Verma, Lou, Liu, Zhang, and Yin}]{ahn2024largelanguagemodelsmathematical}
\bibinfo{author}{J.~Ahn}, \bibinfo{author}{R.~Verma}, \bibinfo{author}{R.~Lou}, \bibinfo{author}{D.~Liu}, \bibinfo{author}{R.~Zhang}, \bibinfo{author}{W.~Yin},
\newblock \bibinfo{title}{Large language models for mathematical reasoning: Progresses and challenges},
\newblock in: \bibinfo{editor}{N.~Falk}, \bibinfo{editor}{S.~Papi}, \bibinfo{editor}{M.~Zhang} (Eds.), \bibinfo{booktitle}{Proceedings of the 18th Conference of the European Chapter of the Association for Computational Linguistics: Student Research Workshop}, \bibinfo{publisher}{Association for Computational Linguistics}, \bibinfo{address}{St. Julian{'}s, Malta}, \bibinfo{year}{2024}, pp. \bibinfo{pages}{225--237}. \URLprefix \url{https://aclanthology.org/2024.eacl-srw.17/}.
\bibitem[{Wei et~al.(2022)Wei, Wang, Schuurmans, Bosma, Ichter, Xia, Chi, Le, and Zhou}]{wei2023chainofthoughtpromptingelicitsreasoning}
\bibinfo{author}{J.~Wei}, \bibinfo{author}{X.~Wang}, \bibinfo{author}{D.~Schuurmans}, \bibinfo{author}{M.~Bosma}, \bibinfo{author}{B.~Ichter}, \bibinfo{author}{F.~Xia}, \bibinfo{author}{E.~Chi}, \bibinfo{author}{Q.~V. Le}, \bibinfo{author}{D.~Zhou},
\newblock \bibinfo{title}{Chain-of-thought prompting elicits reasoning in large language models},
\newblock in: \bibinfo{editor}{S.~Koyejo}, \bibinfo{editor}{S.~Mohamed}, \bibinfo{editor}{A.~Agarwal}, \bibinfo{editor}{D.~Belgrave}, \bibinfo{editor}{K.~Cho}, \bibinfo{editor}{A.~Oh} (Eds.), \bibinfo{booktitle}{Advances in Neural Information Processing Systems}, volume~\bibinfo{volume}{35}, \bibinfo{publisher}{Curran Associates, Inc.}, \bibinfo{year}{2022}, pp. \bibinfo{pages}{24824--24837}. \URLprefix \url{https://proceedings.neurips.cc/paper_files/paper/2022/file/9d5609613524ecf4f15af0f7b31abca4-Paper-Conference.pdf}.
\bibitem[{Chiang et~al.(2024)Chiang, Zheng, Sheng, Angelopoulos, Li, Li, Zhu, Zhang, Jordan, Gonzalez, and Stoica}]{chiang2024chatbot}
\bibinfo{author}{W.-L. Chiang}, \bibinfo{author}{L.~Zheng}, \bibinfo{author}{Y.~Sheng}, \bibinfo{author}{A.~N. Angelopoulos}, \bibinfo{author}{T.~Li}, \bibinfo{author}{D.~Li}, \bibinfo{author}{B.~Zhu}, \bibinfo{author}{H.~Zhang}, \bibinfo{author}{M.~I. Jordan}, \bibinfo{author}{J.~E. Gonzalez}, \bibinfo{author}{I.~Stoica},
\newblock \bibinfo{title}{Chatbot arena: an open platform for evaluating {LLMs} by human preference},
\newblock in: \bibinfo{booktitle}{Proceedings of the 41st International Conference on Machine Learning, {ICML}'24}, \bibinfo{publisher}{JMLR.org}, \bibinfo{year}{2024}.
\bibitem[{Zotos et~al.(2025)Zotos, van Rijn, and Nissim}]{zotos2024doubtful}
\bibinfo{author}{L.~Zotos}, \bibinfo{author}{H.~van Rijn}, \bibinfo{author}{M.~Nissim},
\newblock \bibinfo{title}{Are you doubtful? {O}h, it might be difficult then! {E}xploring the use of model uncertainty for question difficulty estimation},
\newblock in: \bibinfo{editor}{C.~Mills}, \bibinfo{editor}{G.~Alexandron}, \bibinfo{editor}{D.~Taibi}, \bibinfo{editor}{G.~L. Bosco}, \bibinfo{editor}{L.~Paquette} (Eds.), \bibinfo{booktitle}{Proceedings of the 18th International Conference on Educational Data Mining}, \bibinfo{publisher}{International Educational Data Mining Society}, \bibinfo{address}{Palermo, Italy}, \bibinfo{year}{2025}, pp. \bibinfo{pages}{77--89}. \DOIprefix\doi{10.5281/zenodo.15870153}.
\bibitem[{Pezeshkpour and Hruschka(2024)}]{pezeshkpour2023large}
\bibinfo{author}{P.~Pezeshkpour}, \bibinfo{author}{E.~Hruschka},
\newblock \bibinfo{title}{Large language models sensitivity to the order of options in multiple-choice questions},
\newblock in: \bibinfo{editor}{K.~Duh}, \bibinfo{editor}{H.~Gomez}, \bibinfo{editor}{S.~Bethard} (Eds.), \bibinfo{booktitle}{Findings of the Association for Computational Linguistics: {NAACL} 2024}, \bibinfo{publisher}{Association for Computational Linguistics}, \bibinfo{address}{Mexico City, Mexico}, \bibinfo{year}{2024}, pp. \bibinfo{pages}{2006--2017}. \URLprefix \url{https://aclanthology.org/2024.findings-naacl.130/}. \DOIprefix\doi{10.18653/v1/2024.findings-naacl.130}.
\bibitem[{Plaut et~al.(2024)Plaut, Nguyen, and Trinh}]{plaut2024softmax}
\bibinfo{author}{B.~Plaut}, \bibinfo{author}{K.~Nguyen}, \bibinfo{author}{T.~Trinh},
\newblock \bibinfo{title}{Softmax probabilities (mostly) predict large language model correctness on multiple-choice {Q\&A}},
\newblock \bibinfo{journal}{CoRR} \bibinfo{volume}{abs/2402.13213} (\bibinfo{year}{2024}). \URLprefix \url{https://doi.org/10.48550/arXiv.2402.13213}.
\bibitem[{Zotos et~al.(2025)Zotos, van Rijn, and Nissim}]{zotos-etal-2025-model}
\bibinfo{author}{L.~Zotos}, \bibinfo{author}{H.~van Rijn}, \bibinfo{author}{M.~Nissim},
\newblock \bibinfo{title}{Can model uncertainty function as a proxy for multiple-choice question item difficulty?},
\newblock in: \bibinfo{editor}{O.~Rambow}, \bibinfo{editor}{L.~Wanner}, \bibinfo{editor}{M.~Apidianaki}, \bibinfo{editor}{H.~Al-Khalifa}, \bibinfo{editor}{B.~D. Eugenio}, \bibinfo{editor}{S.~Schockaert} (Eds.), \bibinfo{booktitle}{Proceedings of the 31st International Conference on Computational Linguistics}, \bibinfo{publisher}{Association for Computational Linguistics}, \bibinfo{address}{Abu Dhabi, UAE}, \bibinfo{year}{2025}, pp. \bibinfo{pages}{11304--11316}. \URLprefix \url{https://aclanthology.org/2025.coling-main.749/}.

\end{thebibliography}

\end{document}